# The CQC Algorithm: Cycling in Graphs to Semantically Enrich and Enhance a Bilingual Dictionary

**Tiziano Flati**                                                    FLATI@DI.UNIROMA1.IT
**Roberto Navigli**                                                  NAVIGLI@DI.UNIROMA1.IT
*Dipartimento di Informatica, Sapienza University of Rome*
*00198, Rome, Italy.*

## Abstract

Bilingual machine-readable dictionaries are knowledge resources useful in many automatic tasks. However, compared to monolingual computational lexicons like WordNet, bilingual dictionaries typically provide a lower amount of structured information such as lexical and semantic relations, and often do not cover the entire range of possible translations for a word of interest. In this paper we present Cycles and Quasi-Cycles (CQC), a novel algorithm for the automated disambiguation of ambiguous translations in the lexical entries of a bilingual machine-readable dictionary. The dictionary is represented as a graph, and cyclic patterns are sought in this graph to assign an appropriate sense tag to each translation in a lexical entry. Further, we use the algorithm's output to improve the quality of the dictionary itself, by suggesting accurate solutions to structural problems such as misalignments, partial alignments and missing entries. Finally, we successfully apply CQC to the task of synonym extraction.

## 1. Introduction

Lexical knowledge resources, such as thesauri, machine-readable dictionaries, computational lexicons and encyclopedias, have been enjoying increasing popularity over the last few years. Among such resources we cite Roget's Thesaurus (Roget, 1911), the Macquarie Thesaurus (Bernard, 1986), the Longman Dictionary of Contemporary English (Proctor, 1978, LDOCE), WordNet (Fellbaum, 1998) and Wikipedia. These knowledge resources have been utilized in many applications, including Word Sense Disambiguation (Yarowsky, 1992; Nastase & Szpakowicz, 2001; Martínez, de Lacalle, & Agirre, 2008, cf. Navigli, 2009b, 2012 for a survey), semantic interpretation of text (Gabrilovich & Markovitch, 2009), Semantic Information Retrieval (Krovetz & Croft, 1992; Mandala, Tokunaga, & Tanaka, 1998; Sanderson, 2000), Question Answering (Lita, Hunt, & Nyberg, 2004; Moldovan & Novischi, 2002), Information Extraction (Jacquemin, Brun, & Roux, 2002), knowledge acquisition (Navigli & Ponzetto, 2010), text summarization (Silber & McCoy, 2003; Nastase, 2008), classification (Rosso, Molina, Pla, Jiménez, & Vidal, 2004; Wang & Domeniconi, 2008; Navigli, Faralli, Soroa, de Lacalle, & Agirre, 2011) and even simplification (Woodsend & Lapata, 2011).

Most of these applications exploit the structure provided by the adopted lexical resources in a number of different ways. For instance, lexical and semantic relations encoded in computational lexicons such as WordNet have been shown to be very useful in graph-based Word Sense Disambiguation (Mihalcea, 2005; Agirre & Soroa, 2009; Navigli & Lapata, 2010; Ponzetto & Navigli, 2010) and semantic similarity (Pedersen, Banerjee, & Patwardhan, 2005; Agirre, Alfonseca, Hall, Kravalova, Pasca, & Soroa, 2009). Interestingly, it has been reported that the higher the amount of structured knowledge, the higher the disambiguation performance (Navigli & Lapata, 2010; Cuadros & Rigau, 2006). Unfortunately, not all the semantics are made explicit within lexical resources. Even WordNet (Fellbaum, 1998), the most widely-used computational lexicon of English, provides explanatory information in the unstructured form of textual definitions, i.e., strings of text which explain the meaning of concepts using possibly ambiguous words (e.g., "motor vehicle with four wheels" is provided as a definition of the most common sense of *car*). Still worse, while computational lexicons like WordNet contain semantically explicit information such as *is-a* and *part-of* relations, machine-readable dictionaries (MRDs) are often just electronic transcriptions of their paper counterparts. Thus, for each entry they mostly provide implicit information in the form of free text, which cannot be immediately utilized in Natural Language Processing applications.

Over recent years various approaches to the disambiguation of monolingual dictionary definitions have been investigated (Harabagiu, Miller, & Moldovan, 1999; Litkowski, 2004; Castillo, Real, Asterias, & Rigau, 2004; Navigli & Velardi, 2005; Navigli, 2009a), and results have shown that they can, indeed, boost the performance of difficult tasks such as Word Sense Disambiguation (Cuadros & Rigau, 2008; Agirre & Soroa, 2009). However, little attention has been paid to the disambiguation of bilingual dictionaries, which would be capable of improving popular applications such as Machine Translation.

In this article we present a graph-based algorithm which aims at disambiguating translations in bilingual machine-readable dictionaries. Our method takes as input a bilingual MRD and transforms it into a graph whose nodes are word senses[1] (e.g., $car_n^1$) and whose edges $(s, s')$ mainly represent the potential relations between the source sense $s$ of a word $w$ (e.g., $car_n^1$) and the various senses $s'$ of its translations (e.g., $macchina_n^3$). Next, we introduce a novel notion of cyclic and quasi-cyclic graph paths that we use to select the most appropriate sense for a translation $w'$ of a source word $w$.

The contributions of this paper are threefold: first, we present a novel graph-based algorithm for the disambiguation of bilingual dictionaries; second, we exploit the disambiguation results in a way which should help lexicographers make considerable improvements to the dictionary and address issues or mistakes of various kinds; third, we use our algorithm to automatically identify synonyms aligned across languages.

The paper is organized as follows: in Section 2 we introduce the reader to the main ideas behind our algorithm, also with the help of a walk-through example. In Section 3 we provide preliminary definitions needed to introduce our disambiguation algorithm. In Section 4 we present the Cycles and Quasi-Cycles (CQC) algorithm for the disambiguation of bilingual dictionaries. In Section 5 we assess its disambiguation performance on dictionary translations. In Section 6, we show how to enhance the dictionary semi-automatically by means of CQC, and provide experimental evidence in Section 7. In Section 8 we describe an application to monolingual and bilingual synonym extraction and then in Section 9 describe experiments. Related work is presented in Section 10. We give our conclusions in Section 11.

---

1. We denote with $w_p^i$ the $i$-th sense of a word $w$ with part of speech $p$ in a reference sense inventory (we use $n$ for nouns, $v$ for verbs, $a$ for adjectives and $r$ for adverbs), where senses can simply be denoted by integers (like 1, 2, 3, etc.), but also by letters and numbers (such as $A.1$, $B.4$, $D.3$) indicating different levels of granularity (homonymy, polysemy, etc.).

## 2. A Brief Overview

In this section we provide a brief overview of our approach to the disambiguation of bilingual dictionary entries.

### 2.1 Goal

The general form of a bilingual dictionary entry is:

$$w_p^i \rightarrow v_1, v_2, \ldots, v_k$$

where:

- $w_p^i$ is the $i$-th sense of the word $w$ with part of speech $p$ in the source language (e.g., $play_v^2$ is the second sense of the verb $play$);

- each $v_j$ is a translation in the target language for sense $w_p^i$ (e.g., $suonare_v$ is a translation for $play_v^2$). Note that each $v_j$ is implicitly assumed to have the same part of speech $p$ as $w_p$. Importantly, no sense is explicitly associated with $v_j$.

Our objective is to associate each target word $v_j$ with one of its senses so that the concepts expressed by $w_p$ and $v_j$ match. We aim to do this in a systematic and automatic way. First of all, starting from a bilingual dictionary (see Section 3.1), we build a "noisy" graph associated with the dictionary (see Section 3.2), whose nodes are word senses and edges are (mainly) translation relations between word senses. These translation relations are obtained by linking a source word sense ($w_p^i$ above) to all the senses of a target word $v_j$. Next, we define a novel notion of graph patterns, which we have called Cycles and Quasi-Cycles (CQC), that we use as a support for predicting the most suitable sense for each translation $v_j$ of a source word sense $w_p^i$ (see Section 3.3).

### 2.2 A Walk-Through Example

We now present a walk-through example to give further insights into the main goal of the present work. Consider the following Italian-English dictionary entries:

$$
\begin{array}{rcl}
giocare_v^{A.1} & \rightarrow & play,\ toy \\
recitare_v^{A.2} & \rightarrow & act,\ play \\
suonare_v^{A.1} & \rightarrow & sound,\ ring,\ play \\
suonare_v^{B.4} & \rightarrow & ring,\ echo \\
interpretare_v^4 & \rightarrow & play,\ act
\end{array}
$$

and the following English-Italian entries:

$$
\begin{array}{rcl}
play_v^1 & \rightarrow & giocare \\
play_v^2 & \rightarrow & suonare,\ riprodurre \\
play_v^3 & \rightarrow & interpretare,\ recitare
\end{array}
$$

Our aim is to sense tag the target terms on the right-hand side, i.e., we would like to obtain the following output:

$$giocare_v^{A.1} \rightarrow play_v^1,\ toy_v^1$$
$$recitare_v^{A.2} \rightarrow act_v^{A.1},\ play_v^3$$
$$suonare_v^{A.1} \rightarrow sound_v^{1.A.1},\ ring_v^{2.A.2},\ play_v^2$$
$$suonare_v^{B.4} \rightarrow ring_v^{2.A.4},\ echo_v^{A.1}$$
$$interpretare_v^4 \rightarrow play_v^3,\ act_v^{A.1}$$

$$play_v^1 \rightarrow giocare_v^{A.1}$$
$$play_v^2 \rightarrow suonare_v^{A.2},\ riprodurre_v^1$$
$$play_v^3 \rightarrow interpretare_v^3,\ recitare_v^{A.2}$$

where the numbers beside each right-hand translation correspond to the most suitable senses in the dictionary for that translation (e.g., the first sense of $play_v$ corresponds to the sense of playing a game). For instance, in order to disambiguate the first entry above (i.e., $giocare_v^{A.1} \rightarrow play,\ toy$), we have to determine the best sense of the English verb *play* given the Italian verb sense $giocare_v^{A.1}$. We humans know that since the source sense is about "playing a game", the right sense for $play_v$ is the first one. In fact, among the 3 senses of the verb $play_v$ shown above, we can see that the first sense is the only one which translates back into *giocare*. In other words, the first sense of $play_v$ is the only one which is contained in a path starting from, and ending in, $giocare_v^{A.1}$, namely:

$$giocare_v^{A.1} \rightarrow play_v^1 \rightarrow giocare_v^{A.1}$$

while there are no similar paths involving the other senses of $play_v$. Our hunch is that by exploiting cyclic paths we are able to predict the most suitable sense of an ambiguous translation. We provide a scoring function which weights paths according to their length (with shorter paths providing better clues, and thus receiving higher weights) and, at the same time, favours senses which participate in most paths. We will also study the effect of edge reversal as a further support for disambiguating translations. Our hunch here is that by allowing the reversal of subsequent edges we enable previously-missed meaningful paths, which we call quasi-cycles (e.g., $recitare_v^{A.2} \rightarrow play_v^3 \rightarrow interpretare_v^3 \rightarrow act_v^{A.1} \leftarrow recitare_v^{A.2}$). We anticipate that including quasi-cycles significantly boosts the overall disambiguation performance.

## 3. Preliminaries

We now provide some fundamental definitions which will be used throughout the rest of the paper.

### 3.1 Bilingual Dictionary

We define a **bilingual machine-readable dictionary** (BiMRD) as a quadruple $D = (\mathcal{L}, Senses, \mathcal{T}, \mathcal{M})$, where $\mathcal{L}$ is the bilingual lexicon (i.e., $\mathcal{L}$ includes all the lexical items for both languages), $Senses$ is a mapping such that, given a lexical item $w \in \mathcal{L}$, returns the set of senses for $w$ in $D$, $\mathcal{T}$ is a translation function which, given a word sense $s \in Senses(w)$, provides a set of (possibly ambiguous) translations for $s$. Typically, $\mathcal{T}(s) \subset \mathcal{L}$, that is, the translations are in the lexicon. However, it might well be that some translations in $\mathcal{T}(s)$ are not in the lexicon. Finally, $\mathcal{M}$ is a function which, given a word sense $s \in Senses(w)$, provides the set of all words representing meta-information for sense $s$ (e.g., $\mathcal{M}(phoneme_n^1)$ = {$linguistics$}).

| |
|---|
| **language** [ˈlæŋgwidʒ] *n.* |
| **1** lingua; linguaggio: **foreign languages,** lingue straniere; **technical l.,** la lingua della tecnica; **the l. of poetry,** il linguaggio poetico; **dead languages,** le lingue morte ● **l. laboratory,** laboratorio linguistico □ **bad l.,** linguaggio scorretto (*o* sboccato) □ **sign l.,** lingua dei segni (*usata dai sordomuti*) □ **strong l.,** linguaggio violento (*o* volgare) □ **to use bad l.,** usare un linguaggio volgare, da trivio. <br> **2** favella: **Animals do not possess l.,** gli animali non possiedono la favella. |

Figure 1: Entry example of the Ragazzini-Biagi dictionary.

For instance, consider the Ragazzini-Biagi English-Italian BiMRD (Ragazzini & Biagi, 2006). The dictionary provides Italian translations for each English word sense, and vice versa. For a given source lemma (e.g., $language_n$ in English), the dictionary lists its translations in the target language for each sense expressed by the lemma. Figure 1 shows the dictionary entry of $language_n$. The dictionary provides:

- a lexicon for the two languages, i.e., the set $\mathcal{L}$ of lemmas for which dictionary entries exist (such as $language_n$ in Figure 1, but also $lingua_n$, $linguaggio_n$, etc.);

- the set of senses of a given lemma, e.g., $Senses(language_n) = \{language_n^1, language_n^2\}$ (the communication sense vs. the speaking ability), $Senses(lingua_n) = \{lingua_n^1, lingua_n^2\}$ (the muscular organ and a set of words used for communication, respectively); $Senses(linguaggio_n) = \{linguaggio_n^1, linguaggio_n^2, linguaggio_n^3\}$ (the faculty of speaking, the means of communication and machine language, respectively);

- the translations for a given sense, e.g., $\mathcal{T}(language_n^1) = \{lingua_n, linguaggio_n\}$;

- optionally, some meta-information about a given sense, such as $\mathcal{M}(phoneme_n^1) = \{linguistics\}$.

The dictionary also provides usage examples and compound translations (see Figure 1), lexical variants (e.g., *acknowledgement* vs. *acknowledgment*) and references to other entries (e.g., from *motorcar* to *car*).

## 3.2 Noisy Graph

Given a BiMRD $D$, we define a **noisy dictionary graph** $G = (V, E)$ as a directed graph where:

1. $V$ is the set of senses in the dictionary $D$ (i.e., $V = \bigcup_{w \in \mathcal{L}} Senses(w)$);

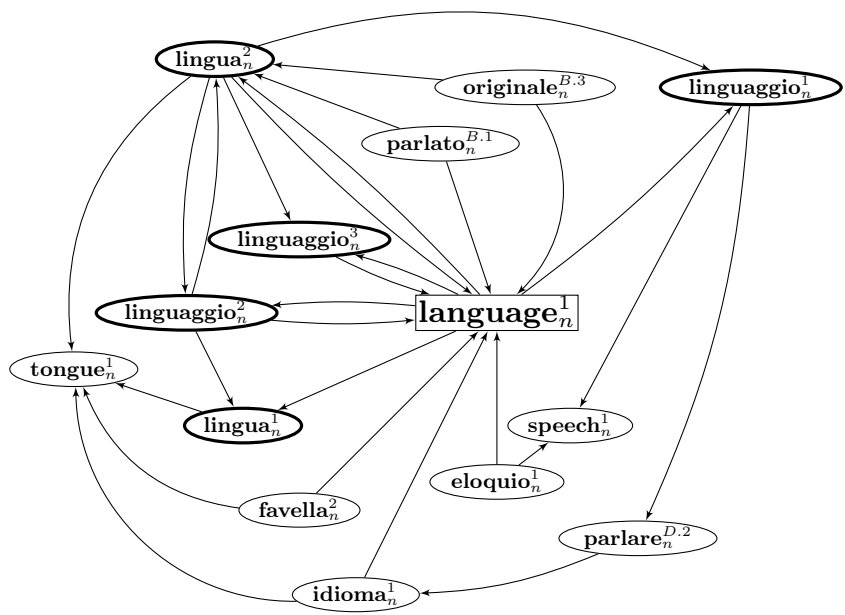

Figure 2: An excerpt of the Ragazzini-Biagi noisy graph including $language_n^1$ and its neighbours.

2. For each word $w \in \mathcal{L}$ and a sense $s \in Senses(w)$, an edge $(s, s')$ is in $E$ if and only if $s'$ is a sense of a translation of $s$ in the dictionary (i.e., $s' \in Senses(w')$ and $w' \in \mathcal{T}(s)$), or $s'$ is a sense of a meta-word $m$ in the definition of $s$ (i.e., if $s' \in Senses(m)$ for some $m \in \mathcal{M}(s)$).

According to the above definition, given an ambiguous word $w'$ in the definition of $s$, we add an edge from $s$ to each sense of $w'$ in the dictionary. In other words, the noisy graph $G$ associated with dictionary $D$ encodes all the potential meanings for word translations in terms of edge connections. In Figure 2 we show an excerpt of the noisy graph associated with the Ragazzini-Biagi dictionary. In this sub-graph three kinds of nodes can be found:

- the source sense (rectangular box), namely $language_n^1$.

- the senses of its translations (thick ellipse-shaped nodes), e.g., the three senses of $linguaggio_n$ and the two senses of $lingua_n$.

- other senses (ellipse-shaped nodes), which are either translations or meta-information for other senses (e.g., $speech_n^1$ is a translation sense of $eloquio_n^1$).

## 3.3 Graph Cycles and Quasi-Cycles

We now recall the definition of **graph cycle**. A cycle for a graph $G$ is a sequence of edges of $G$ that form a path $v_1 \to v_2 \to \cdots \to v_n$ ($v_i \in V \ \forall i \in \{1, \ldots, n\}$) such that the first node

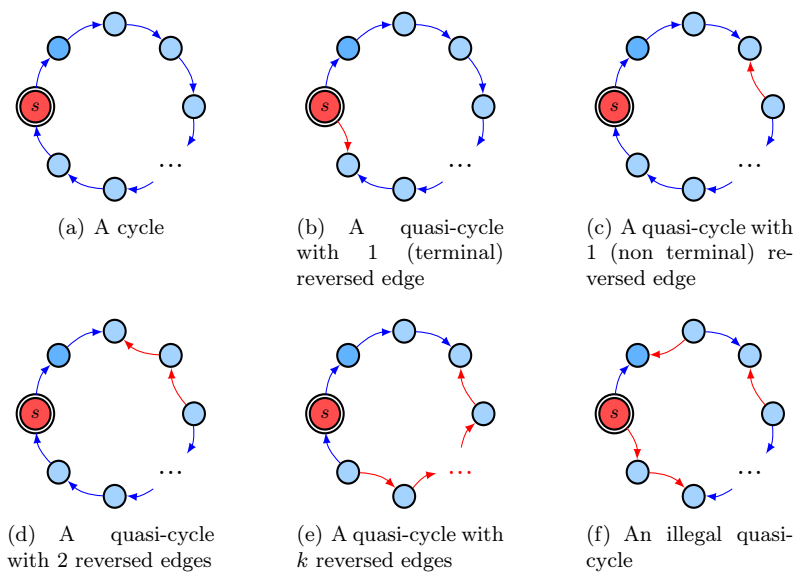

(a) A cycle

(b) A quasi-cycle with 1 (terminal) reversed edge

(c) A quasi-cycle with 1 (non terminal) reversed edge

(d) A quasi-cycle with 2 reversed edges

(e) A quasi-cycle with $k$ reversed edges

(f) An illegal quasi-cycle

Figure 3: Legal and illegal cycles and quasi-cycles.

of the path corresponds to the last, i.e., $v_1 = v_n$ (Cormen, Leiserson, & Rivest, 1990, p. 88). The length of a cycle is given by the number of its edges. For example, a cycle of length 3 in Figure 2 is given by the path:

$$language_n^1 \rightarrow linguaggio_n^2 \rightarrow lingua_n^2 \rightarrow language_n^1.$$

We further provide the definition of **quasi-cycle** as a sequence of edges in which the reversal of the orientation of one or more consecutive edges creates a cycle (Bohman & Thoma, 2000). For instance, a quasi-cycle of length 4 in Figure 2 is given by the path:

$$language_n^1 \rightarrow linguaggio_n^1 \rightarrow speech_n^1 \leftarrow eloquio_n^1 \rightarrow language_n^1.$$

It can be seen that the reversal of the edge $(eloquio_n^1, speech_n^1)$ creates a cycle. Since the direction of this edge is opposite to that of the cycle, we call it a *reversed edge*. Finally, we say that a path is (quasi-)cyclic if it forms a (quasi-)cycle. Note that we do not consider paths going across senses of the same word; so $language_n^1 \rightarrow \underline{lingua_n^1} \rightarrow tongue_n^1 \leftarrow \underline{lingua_n^2} \rightarrow language_n^1$ is not considered a legal quasi-cycle.

In order to provide a graphical representation of (quasi-)cycles, in Figure 3 we show different kinds of (quasi-)cycles starting from a given node $s$, namely: a cycle (a), a quasi-cycle with 1 terminal (b) and non-terminal (c) reversed edge (a reversed edge is terminal if it is incident from $s$), with more reversed edges ((d) and (e)), and an illegal quasi-cycle whose reversed edges are not consecutive (f).

## 4. The CQC Algorithm

We are now ready to introduce the Cycles & Quasi-Cycles (CQC) algorithm, whose pseudocode is given in Table 1. The algorithm takes as input a BiMRD $D = (\mathcal{L}, Senses, \mathcal{T}, \mathcal{M})$, and a sense $s$ of a word $w$ in its lexicon (i.e., $w \in \mathcal{L}$ and $s \in Senses(w)$). The algorithm aims at disambiguating each of the word's ambiguous translations $w' \in \mathcal{T}(s)$, i.e., to assign it the right sense among those listed in $Senses(w')$.

| | **CQC**(BiMRD $D = (\mathcal{L}, Senses, \mathcal{T}, \mathcal{M})$, sense $s$ of $w \in \mathcal{L}$) |
|---|---|
| 1 | **for each** word $w' \in \mathcal{T}(s)$ |
| 2 | **for each** sense $s' \in Senses(w')$ |
| 3 | $paths(s') := \text{DFS}(s', s)$ |
| 4 | $all\_paths := \bigcup_{s' \in Senses(w')} paths(s')$ |
| 5 | **for each** sense $s' \in Senses(w')$ |
| 6 | $score(s') := 0$ |
| 7 | **for each** path $p \in paths(s')$ |
| 8 | $l := length(p)$ |
| 9 | $v := \omega(l) \cdot \frac{1}{NumPaths(all\_paths, l)}$ |
| 10 | $score(s') := score(s') + v$ |
| 11 | $\mu(w') = \underset{s' \in Senses(w')}{\operatorname{argmax}}\ score(s')$ |
| 12 | **return** $\mu$ |

Table 1: The Cycles & Quasi-Cycles (CQC) algorithm in pseudocode.

The algorithm outputs a mapping $\mu$ between each ambiguous word $w' \in \mathcal{T}(s)$ and the sense $s'$ of $w'$ chosen as a result of the disambiguation procedure that we illustrate hereafter.

First, for each sense $s'$ of our target translation $w' \in \mathcal{T}(s)$, the algorithm performs a search of the noisy graph associated with $D$ and collects the following kinds of paths:

i) Cycles:

$$s \to s' \to s_1 \to \cdots \to s_{n-2} \to s_{n-1} = s$$

ii) Quasi-cycles:

$$s \to s' \to s_1 \to ... \to s_j \leftarrow ... \leftarrow s_k \to ... \to s_{n-2} \to s_{n-1} = s \qquad 1 \le j \le n-2, j < k \le n-1 \tag{1}$$

where $s$ is our source sense, $s'$ is our candidate sense for $w' \in \mathcal{T}(s)$, $s_i$ is a sense listed in $D$ ($i \in \{1, \ldots, n-2\}$), $s_{n-1} = s$, and $n$ is the length of the path. Note that both kinds of path start and end with the same node $s$, and that the algorithm searches for quasi-cycles whose reversed edges connecting $s_k$ to $s_j$ are consecutive. To avoid redundancy we require (quasi-)cycles to be simple, that is, no node is repeated in the path except the start/end node (i.e., $s_i \ne s, s_i \ne s', s_i \ne s_{i'}\ \forall i, i'$ s. t. $i \ne i'$).

During the first step of the algorithm (see Table 1, lines 2-3), (quasi-)cyclic paths are sought for each sense of $w'$. This step is performed with a depth-first search (DFS, cf. Cormen et al., 1990, pp. 477–479) up to a depth $\delta$.[2] The DFS – whose pseudocode is shown in Table 2 – starts from a sense $s' \in Senses(w')$, and recursively explores the graph; outgoing edges are explored in order to collect cycles (lines 7-9 of Rec-DFS, see Table 2) while incoming edges are considered in order to collect quasi-cycles (lines 11-14); before extending the current path $p$ with a reversed edge, however, it is necessary to check whether the latter is consecutive to all previously reversed edges (if any) present in $p$ and to skip it otherwise (cf. Formula (1)). The stack $visited$ contains the nodes visited so far, in order to avoid the repetition of a node in a path (cf. lines 1, 5 and 15 of Rec-DFS). Finally the search ends when the maximum path length is reached, or a previously visited node is encountered (line 1 of Rec-DFS); otherwise, if the initial sense $s$ is found, a (quasi-)cycle is collected (lines 2-4 of Rec-DFS). For each sense $s'$ of $w'$ the DFS returns the full set $paths(s')$ of paths collected. Finally, in line 4 of Table 1, $all\_paths$ is set to store the paths for all the senses of $w'$.

---

2. We note that a depth-first search is equivalent to a breadth-first search (BFS) for the purpose of collecting paths.

| | **DFS**(sense $s'$, sense $s$) |
|---|---|
| 1 | $paths := \emptyset$ |
| 2 | $visited := \emptyset$ |
| 3 | Rec-DFS($s'$, $s$, $s \to s'$) |
| 4 | **return** paths |

| | **Rec-DFS**(sense $s'$, sense $s$, path $p$) |
|---|---|
| 1 | **if** $s' \in visited$ **or** $length(p) > \delta$ **then return** |
| 2 | **if** $s' = s$ **then** |
| 3 | $paths := paths \cup \{p\}$ |
| 4 | **return** |
| 5 | $push(visited, s')$ |
| 6 | // cycles |
| 7 | **for each** edge $s' \to s''$ |
| 8 | $p' := p \to s''$ |
| 9 | Rec-DFS($s''$, $s$, $p'$) |
| 10 | // quasi-cycles |
| 11 | **for each** edge $s' \leftarrow s''$ |
| 12 | $p' := p \leftarrow s''$ |
| 13 | **if** reversedEdgesNotConsecutive($p'$) **then continue** |
| 14 | Rec-DFS($s''$, $s$, $p'$) |
| 15 | $pop(visited)$ |

Table 2: The depth-first search pseudocode algorithm for cycle and quasi-cycle collection.

The second phase of the CQC algorithm (lines 5-10 of Table 1) computes a score for each sense $s'$ of $w'$ based on the paths collected for $s'$ during the first phase. Let $p$ be such a path, and let $l$ be its length, i.e., the number of edges in the path. Then the contribution of $p$ to the score of $s'$ is given by:

$$score(p) := \frac{\omega(l)}{NumPaths(all\_paths, l)} \qquad (2)$$

where:

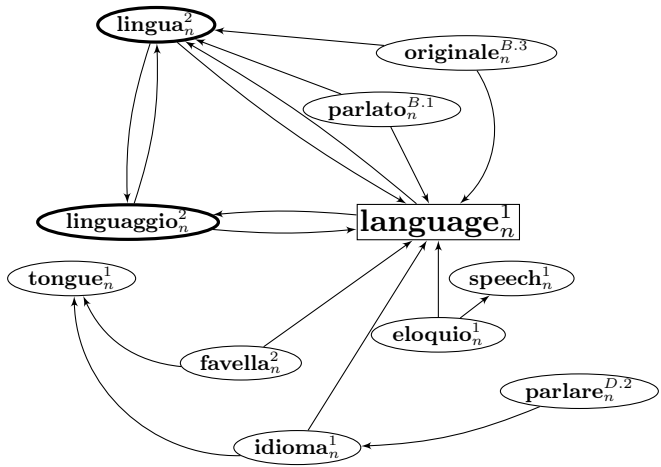

Figure 4: The Ragazzini-Biagi graph from Figure 2 pruned as a result of the CQC algorithm.

- $\omega(l)$ is a monotonically non-increasing function of its length $l$; in our experiments, we tested three different weight functions $\omega(l)$, namely a constant, a linear and an inversely exponential function (see Section 5).

- the normalization factor $NumPaths(all\_paths, l)$ calculates the overall number of collected paths of length $l$ among all the target senses.

In this way the score of a sense $s'$ amounts to:

$$score(s') := \sum_{p \in paths(s')} score(p) = \sum_{l=2}^{\delta} \omega(l) \frac{NumPaths(paths(s'), l)}{NumPaths(all\_paths, l)} \qquad (3)$$

The rationale behind our scoring formula is two-fold: first – thanks to function $\omega$ – it favours shorter paths, which are intuitively less likely to be noisy; second, for each path length, it accounts for the ratio of paths of that length in which $s'$ participates (second factor of the right-hand side of the formula above).

After the scores for each sense $s'$ of the target translation $w'$ have been calculated, a mapping is established between $w'$ and the highest-scoring sense (line 11). Finally, after all the translations have been disambiguated, the mapping is returned (line 12).

As a result of the systematic application of the algorithm to each sense in our BiMRD $D$, a new graph $G' = (V, E')$ is output, where $V$ is again the sense inventory of $D$, and $E'$ is a subset of the noisy edge set $E$ such that each edge $(s, s') \in E'$ is the result of our disambiguation algorithm run with input $D$ and $s$. Figure 4 shows the "clean", unambiguous dictionary graph after executing CQC, as compared to the initial noisy graph from Figure 2. In this pruned graph, each sense links to only one sense of each of its translations.

## 4.1 An Example

As an example, consider the following dictionary entry in the Ragazzini-Biagi dictionary:

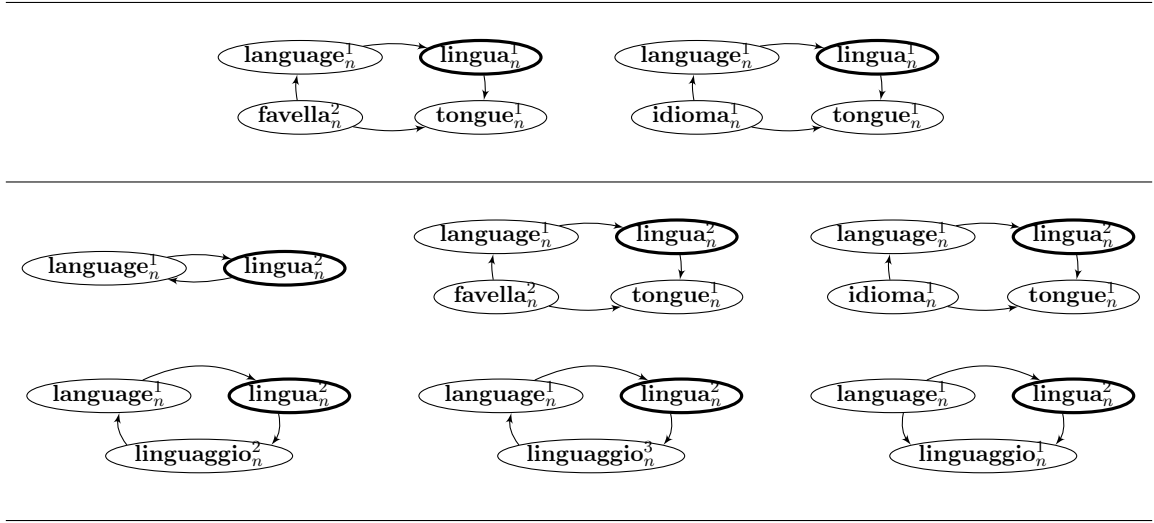

Figure 5: Cycles and quasi-cycles collected from DFS($lingua_n^1$, $language_n^1$) (top) and DFS($lingua_n^2$, $language_n^1$) (bottom).

**language** *n.* **1** lingua; linguaggio.

In order to disambiguate the Italian translations we call the CQC algorithm as follows: CQC($D$, $language_n^1$). Let us first concentrate on the disambiguation of $lingua_n$, an ambiguous word with two senses in the Ragazzini-Biagi. First, two calls are made, namely $DFS(lingua_n^1, language_n^1)$ and $DFS(lingua_n^2, language_n^1)$. Each function call performs a DFS starting from the respective sense of our target word to collect all relevant cycles and quasi-cycles according to the algorithm in Table 2. The set of cycles and quasi-cycles collected for the two senses from the noisy graph of Figure 2 are shown in Figure 5.

During the second phase of the CQC algorithm, and for each sense of $lingua_n$, the contribution of each path is calculated (lines 8-10 of the algorithm in Table 1). Specifically, the following scores are calculated for the two senses of $lingua_n$ (we assume our weight function $\omega(l) = 1/e^l$):

$$score(lingua_n^1) = \quad 2 \ \cdot \ \tfrac{1}{e^4} \ \cdot \ \tfrac{1}{NumPaths(all\_paths, 4)} \simeq 2 \ \cdot 0.018 \ \cdot \tfrac{1}{4} = 0.009$$

$$\begin{aligned} score(lingua_n^2) = \quad & 1 \ \cdot \ \tfrac{1}{e^2} \ \cdot \ \tfrac{1}{NumPaths(all\_paths, 2)} \ + \\ + \ & 3 \ \cdot \ \tfrac{1}{e^3} \ \cdot \ \tfrac{1}{NumPaths(all\_paths, 3)} \ + \\ + \ & 2 \ \cdot \ \tfrac{1}{e^4} \ \cdot \ \tfrac{1}{NumPaths(all\_paths, 4)} \simeq \\ \simeq \quad & 1 \ \cdot 0.135 \ \cdot \tfrac{1}{1} + 3 \ \cdot 0.050 \ \cdot \tfrac{1}{3} + 2 \ \cdot 0.018 \ \cdot \tfrac{1}{4} = 0.194 \end{aligned}$$

where $NumPaths(all\_paths, l)$ is the total number of paths of length $l$ collected over all the senses of $lingua_n$. Finally, the sense with the highest score (i.e., $lingua_n^2$ in our example) is returned.

Similarly, we determine the scores of the various senses of $linguaggio_n$ as follows:

$$score(linguaggio_n^1) = 2 \cdot \frac{1}{e^4} \cdot \frac{1}{NumPaths(all\_paths,4)} \simeq 2 \cdot 0.018 \cdot \frac{1}{4} = 0.009.$$

$$\begin{aligned}
score(linguaggio_n^2) = & 1 \cdot \frac{1}{e^2} \cdot \frac{1}{NumPaths(all\_paths,2)} + \\
+ & 2 \cdot \frac{1}{e^3} \cdot \frac{1}{NumPaths(all\_paths,3)} + \\
+ & 2 \cdot \frac{1}{e^4} \cdot \frac{1}{NumPaths(all\_paths,4)} \simeq \\
\simeq & 1 \cdot 0.135 \cdot \frac{1}{2} + 2 \cdot 0.050 \cdot \frac{1}{2} + 2 \cdot 0.018 \cdot \frac{1}{4} = 0.1265.
\end{aligned}$$

$$score(linguaggio_n^3) = 1 \cdot \frac{1}{e^2} \cdot \frac{1}{NumPaths(all\_paths,2)} \simeq 1 \cdot 0.135 \cdot \frac{1}{2} = 0.0675.$$

As a result, sense 2 is correctly selected.

## 5. Evaluation: Dictionary Disambiguation

In our first set of experiments we aim to assess the disambiguation quality of the CQC algorithm and compare it with existing disambiguation approaches. We first describe our experimental setup in Section 5.1, by introducing the bilingual dictionary used throughout this article, and providing information on the dictionary graph, our tuning and test datasets, and the algorithms, parameters and baselines used in our experiments. We describe our experimental results in Section 5.2.

### 5.1 Experimental Setup

In this section we discuss the experimental setup for our dictionary disambiguation experiment.

#### 5.1.1 DICTIONARY

We performed our dictionary disambiguation experiments on the Ragazzini-Biagi (Ragazzini & Biagi, 2006), a popular bilingual English-Italian dictionary, which contains over 90,000 lemmas and 150,000 word senses.

#### 5.1.2 DICTIONARY GRAPH

In order to get an idea of the difficulty of our dictionary disambiguation task we determined the ratio of wrong edges in the graph. To do this we first calculated the ratio of correct edges, i.e., those edges which link source senses to their right translation senses. This quantity can be estimated as the overall number of translations in the dictionary (i.e., assuming each translation has an appropriate sense in the dictionary) divided by the total number of edges:

$$CorrectnessRatio(G) = \frac{\sum\limits_{s \in V} |\mathcal{T}(s)|}{|E|} \tag{4}$$

The ratio of wrong edges is then calculated as $1 - CorrectnessRatio(G)$, obtaining an estimate of 66.4% of incorrect edges in the noisy graph of the Ragazzini-Biagi dictionary.

| Dataset | # entries | # translations | # polysemous | avg. polysemy | perfect alignments |
|---|---|---|---|---|---|
| Tuning Dataset | 50 | 80 | 53 | 4.74 | 37 |
| Test Dataset | 500 | 1,069 | 765 | 3.95 | 739 |

Table 3: Statistics for the tuning and test datasets.

### 5.1.3 Dataset

Our datasets for tuning and test consist of dictionary entries, each containing translations of a source sense into a target language. Each translation item was manually disambiguated according to its sense inventory in the bilingual dictionary. For example, given the Italian entry $brillante_a^{A.2}$, translated as "$sparkling_a$, $vivacious_a$", we associated the appropriate English sense from the English-Italian section to $sparkling_a$ and $vivacious_a$ (senses 3 and 1, respectively).

For tuning purposes, we created a dataset of 50 entries, totaling 80 translations. We also prepared a test dataset of 500 entries, randomly sampled from the Ragazzini-Biagi dictionary (250 from the English-Italian section, 250 from the Italian-English section). Overall, the test dataset included 1,069 translations to be disambiguated. We report statistics for the two datasets in Table 3, including the number of polysemous translations and the average polysemy of each translation. We note that for 44 of the translations in the test set (i.e., 4.1% of the total) none of the senses listed in the dictionary is appropriate (including monosemous translations). A successful disambiguation system, therefore, should not disambiguate these items. The last column in the table shows the number of translations for which a sense exists that translates back to the source lemma (e.g., $car_n^1$ translates to $macchina$ and $macchina_n^3$ translates to $car$).

### 5.1.4 Algorithms

We compared the following algorithms in our experimental framework[3], since (with the exception of CQC and variants thereof) they represent the most widespread graph-based approaches and are used in many NLP tasks with state-of-the-art performance:

- **CQC**: we applied the CQC algorithm as described in Section 4;

- **Cycles**, a variant of the CQC algorithm which searches for cycles only (i.e., quasi-cycles are not collected);

- **DFS**, which applies an ordinary DFS algorithm and collects all paths between $s$ and $s'$ (i.e., paths are not "closed" by completing them with edge sequences connecting $s'$ to $s$). In this setting the path $s \rightarrow s'$ is discarded, as by construction it can be found in $G$ for each sense $s' \in Senses(w')$;

- **Random walks**, which performs a large number of random walks starting from $s'$ and collecting those paths that lead to $s$. This approach has been successfully used to approximate an exhaustive search of translation circuits (Mausam, Soderland, Etzioni,

---

3. In order to ensure a level playing field, we provided in-house implementations for all the algorithms within our graph-based framework, except for Personalized PageRank, for which we used a standard implementation (http://jung.sourceforge.net).

Weld, Skinner, & Bilmes, 2009; Mausam, Soderland, Etzioni, Weld, Reiter, Skinner, Sammer, & Bilmes, 2010). We note that, by virtue of its simulation nature, this method merely serves as a way of collecting paths at random. In fact, given a path ending in a node $v$, the next edge is chosen equiprobably among all edges outgoing from $v$.

- **Markov chains**, which calculates the probability of arriving at a certain source sense $s$ starting from the initial translation sense $s'$ averaged over $n$ consecutive steps, that is, $p_{s',s} = \frac{1}{n} \sum_{m=1}^{n} p_{s',s}^{(m)}$, where $p_{s',s}^{(m)}$ is the probability of arriving at node $s$ using exactly $m$ steps starting from node $s'$. The initial Markov chain is initialized from the noisy dictionary graph as follows: for each $v, v' \in V$, if $(v, v') \in E$, then $p_{v,v'}^{(0)} = 1/out(v)$, where $out(v)$ is the outdegree of $v$ in the noisy graph, otherwise $p_{v,v'}^{(0)} = 0$.

- **Personalized PageRank (PPR)**: a popular variant of the PageRank algorithm (Brin & Page, 1998) where the original Markov chain approach to node ranking is modified by perturbating the initial probability distribution on nodes (Haveliwala, 2002, 2003). PPR has been successfully applied to Word Sense Disambiguation (Agirre & Soroa, 2009) and thus represents a very competitive system to compare with. In order to disambiguate a target translation $w'$ of a source word $w$, for each translation sense $s'$, we concentrate all the probability mass on $s'$, and apply PPR. We select the best translation sense as the one which maximizes the PPR value of the source word (or, equivalently, that of the translation sense itself).

- **Lesk algorithm** (Lesk, 1986): we apply an adaptation of the Lesk algorithm in which, given a source sense $s$ of word $w$ and a word $w'$ occurring as a translation of $s$, we determine the right sense of $w'$ on the basis of the (normalized) maximum overlap between the entries of each sense $s'$ of $w'$ and that of $s$:

$$\underset{s' \in Senses(w')}{\operatorname{argmax}} \frac{|next^*(s) \cap next^*(s')|}{\max\{|next^*(s)|, |next^*(s')|\}},$$

where we define $next^*(s) = synonyms(s) \cup next(s)$, $synonyms(s)$ is the set of lexicalizations of sense $s$ (i.e., the synonyms of sense $s$, e.g., *acknowledgement* vs *acknowledgment*) and $next(s)$ is the set of nodes $s'$ connected through an edge $(s, s')$.

For all the algorithms that explicitly collect paths (CQC, Cycles, DFS and Random walks), we tried three different functions for weighting paths, namely:

- A constant function $\omega(l) = 1$ that weights all paths equally, independently of their length $l$;

- A linear function $\omega(l) = 1/l$ that assigns each path a score inversely proportional to its length $l$;

- An exponential function $\omega(l) = 1/e^l$ that assigns a score that decreases exponentially with the path length.

| Algorithm | Best Configuration | |
| --- | --- | --- |
| | Length $\delta$ | Specific parameters |
| CQC | 4 | up to 2 terminal reversed edges |
| Cycles | 4 | - |
| DFS | 4 | - |
| Random walks | 4 | 400 random walks |
| Markov chains | 2 | - |

Table 4: Parameter tuning for path-based algorithms.

### 5.1.5 Parameters

We used the tuning dataset to fix the parameters of each algorithm that maximized the performance. We tuned the maximum path length for each of the path-based algorithms (CQC, Cycles, DFS, Random walks and Markov chains), by trying all lengths in $\{1, \ldots, 6\}$. Additionally, for CQC, we tuned the minimum and maximum values for the parameters $j$ and $k$ used for quasi-cyclic patterns (cf. Formula 1 in Section 4). These parameters determine the position and the number of reversed edges in a quasi-cyclic graph pattern. The best results were obtained when $n-1 \leq k \leq n-1$, i.e. $k = n-1$, and $n-3 \leq j < n-1$, that is, CQC yielded the best performance when up to 2 terminal reversed edges were sought (cf. Section 3.3 and Figure 3). For Random walks, we tuned the number of walks needed to disambiguate each item (ranging between 50 and 2,000). The best parameters resulting from tuning are reported in Table 4. Finally, for PPR we used standard parameters: we performed 30 iterations and set the damping factor to 0.85.

### 5.1.6 Measures

To assess the performance of our algorithms, we calculated precision (the number of correct answers over the number of items disambiguated by the system), recall (the number of correct answers over the number of items in the dataset), and F1 (a harmonic mean of precision and recall, given by $\frac{2PR}{P+R}$). Note that precision and recall do not consider those items in the test set for which no appropriate sense is available in the dictionary. In order to account for these items, we also calculated accuracy as the number of correct answers divided by the total number of items in the test set.

### 5.1.7 Baselines

We compared the performance of our algorithms with three baselines:

- the **First Sense (FS) Baseline**, that associates the first sense listed by the dictionary with each translation to be disambiguated (e.g., $car_n^1$ is chosen for $car$ independently of the disambiguation context). The rationale behind this baseline derives from the tendency of lexicographers to sort senses according to the importance they perceive or estimate from a (possibly sense-tagged) corpus;

- the **Random Baseline**, which selects a random sense for each target translation;

- the **Degree Baseline**, that chooses the translation sense with the highest out-degree, i.e., the highest number of outgoing edges.

## 5.2 Results

We are now ready to present the results of our dictionary disambiguation experiment.

### 5.2.1 RESULTS WITHOUT BACKOFF STRATEGY

In Table 5 we report the results of our algorithms on the test set. CQC, PPR and Cycles are the best performing algorithms, achieving around 83%, 81% and 75% accuracy respectively. CQC outperforms all other systems in terms of F1 by a large margin. The results show that the mere use of cyclic patterns does not lead to state-of-the-art performance, which is, instead, obtained when quasi-cycles are also considered. Including quasi-cycles leads to a considerable increase in recall, while at the same time maintaining a high level of precision. The DFS is even more penalizing because it does not get backward support as happens for cycling patterns. Markov chains consistently outperform Random walks. We hypothesize that this is due to the higher coverage of Markov chains compared to the number of random walks collected by a simulated approach. PPR considerably outperforms the two other probabilistic approaches (especially in terms of recall and accuracy), but lags behind CQC by 3 points in F1 and 2 in accuracy. This result confirms previous findings in the literature concerning the high performance of PPR, but also corroborates our hunch about quasi-cycles being the determining factor in the detection of hard-to-find semantic connections within dictionaries. Finally, Lesk achieves high precision, but low recall and accuracy, due to the lack of a lookahead mechanism.

The choice of the weighting function impacts the performance of all path-based algorithms, with $1/e^l$ performing best and the constant function 1 resulting in the worst results (this is not the case for the DFS, though).

The random baseline represents our lowerbound and is much lower than all other results. Compared to the first sense baseline, CQC, PPR and Cycles obtain better performance. This result is consistent with previous findings for tasks such as the Senseval-3 Gloss Word Sense Disambiguation (Litkowski, 2004). However, at the same time, it is in contrast with results on all-words Word Sense Disambiguation (Navigli, 2009b), where the first or most frequent sense baseline generally outperforms most disambiguation systems. Nevertheless, the nature of these two tasks is very different, because – in dictionary entries – senses tend to be equally distributed, whereas in open text they have a single predominant meaning that is determined by context. As for the Degree Baseline, it yields results below expectations, and far worse than the FS baseline. The reason behind this lies in the fact that the amount of translations and translation senses does not necessarily correlate with mainstream meanings.

While attaining the highest precision, CQC also outperforms the other algorithms in terms of accuracy. However, accuracy is lower than F1: this is due to F1 being a harmonic mean of precision and recall, while in calculating accuracy each and every item in the dataset is taken into account, even those items for which no appropriate sense tag can be given.

In order to verify the reliability of our tuning phase (see Section 5.1), we studied the F1 performance of CQC by varying the depth $\delta$ of the DFS (cf. Section 4). The best results – shown in Table 6 – are obtained on the test set when $\delta = 4$, which confirms this as the optimal parameter choice for CQC (cf. Table 4). In fact, F1 increases with higher values of $\delta$, up to a performance peak of 85.19% obtained when $\delta = 4$. With higher values of $\delta$ we observed a performance decay due to the noise introduced. The optimal value of $\delta$ is in line with previous experimental results on the impact of the DFS depth in Word Sense Disambiguation (Navigli & Lapata, 2010).

| Algorithm | Performance | | | |
|---|---|---|---|---|
| | P | R | F1 | A |
| CQC $1/e^l$ | 87.14 | 83.32 | 85.19 | 83.35 |
| CQC $1/l$ | 87.04 | 83.22 | 85.09 | 83.26 |
| CQC 1 | 86.33 | 82.54 | 84.39 | 82.60 |
| Cycles $1/e^l$ | 87.17 | 74.93 | 80.59 | 75.58 |
| Cycles $1/l$ | 86.49 | 74.34 | 79.96 | 75.02 |
| Cycles 1 | 84.56 | 72.68 | 78.17 | 73.43 |
| DFS $1/e^l$ | 63.40 | 37.85 | 47.40 | 39.85 |
| DFS $1/l$ | 63.40 | 37.85 | 47.40 | 39.85 |
| DFS 1 | 63.56 | 37.95 | 47.52 | 39.94 |
| Random walks $1/e^l$ | 83.94 | 61.17 | 70.77 | 62.49 |
| Random walks $1/l$ | 83.67 | 60.98 | 70.55 | 62.30 |
| Random walks 1 | 79.38 | 57.85 | 66.93 | 59.31 |
| Markov chains | 85.46 | 65.37 | 74.08 | 66.70 |
| PPR | 83.20 | 81.25 | 82.21 | 81.27 |
| Lesk | 86.05 | 31.90 | 46.55 | 34.52 |
| First Sense BL | 72.67 | 73.17 | 72.92 | 73.53 |
| Random BL | 28.53 | 29.76 | 29.13 | 28.53 |
| Degree BL | 58.39 | 58.85 | 58.39 | 58.62 |

Table 5: Disambiguation performance on the Ragazzini-Biagi dataset.

| | $\delta = 2$ | $\delta = 3$ | $\delta = 4$ | $\delta = 5$ |
|---|---|---|---|---|
| CQC-$\frac{1}{e^l}$ | 76.94 | 82.85 | 85.19 | 84.50 |

Table 6: Disambiguation performance of CQC-$\frac{1}{e^l}$ based on F1.

### 5.2.2 Results with Backoff Strategy

As mentioned above, the experimented path-based approaches are allowed not to return any result; this is the case when no paths can be found for any sense of the target word. In a second set of experiments we thus let the algorithms use the first sense baseline as a backoff strategy whenever they were not able to give any result for a target word. This is especially useful when the disambiguation system cannot make any decision because of lack of knowledge in the dictionary graph. As can be seen in Table 7, the scenario changes radically in this setting. The adoption of the first sense backoff strategy results in generally higher performance; notwithstanding this CQC keeps showing the best results, achieving almost 87% F1 and accuracy when $\omega = 1/e^l$.

| Algorithm | PERFORMANCE WITH FS | | | |
|---|---|---|---|---|
| | P | R | F1 | A |
| CQC $1/e^l$ | 86.52 | 87.02 | 86.77 | 86.81 |
| CQC $1/l$ | 86.42 | 86.93 | 86.67 | 86.72 |
| CQC 1 | 85.74 | 86.24 | 86.00 | 86.06 |
| Cycles $1/e^l$ | 85.55 | 86.05 | 85.80 | 85.87 |
| Cycles $1/l$ | 84.97 | 85.46 | 85.21 | 85.31 |
| Cycles 1 | 83.32 | 83.80 | 83.56 | 83.72 |
| DFS $1/e^l$ | 68.00 | 68.39 | 68.19 | 68.94 |
| DFS $1/l$ | 68.00 | 68.39 | 68.19 | 68.94 |
| DFS 1 | 68.09 | 68.49 | 68.29 | 69.04 |
| Random walks $1/e^l$ | 82.06 | 82.54 | 82.30 | 82.51 |
| Random walks $1/l$ | 81.86 | 82.34 | 82.10 | 82.32 |
| Random walks 1 | 78.76 | 79.22 | 79.00 | 79.33 |
| Markov chains | 82.75 | 83.32 | 83.03 | 83.26 |
| PPR | 83.12 | 83.77 | 83.44 | 83.12 |
| Lesk | 82.07 | 82.63 | 82.35 | 82.60 |
| First Sense BL | 72.67 | 73.17 | 72.92 | 73.53 |
| Random BL | 28.53 | 29.76 | 29.13 | 28.53 |
| Degree BL | 58.39 | 58.85 | 58.39 | 58.62 |

Table 7: Disambiguation performance on the Ragazzini-Biagi dataset using the first sense (FS) as backoff strategy.

### 5.2.3 DIRECTED VS. UNDIRECTED SETTING

Our algorithm crucially takes advantage of the quasi-cyclic pattern and therefore relies heavily on the directionality of edges. Thus, in order to further verify the beneficial impact of quasi-cycles, we also compared our approach in an undirected setting, i.e., using a noisy graph whose edges are unordered pairs. This setting is similar to that of de Melo and Weikum (2010), who aim at detecting imprecise or wrong interlanguage links in Wikipedia. However, in their task only few edges are wrong (in fact, they remove less than 2% of the cross-lingual interlanguage links), whereas our dictionary graph contains much more noise, which we estimated to involve around 66% of the edges (see Section 5.1.2).

To test whether directionality really matters, we compared CQC with its natural undirected counterpart, namely **Undirected Cycles**: this algorithm collects all (undirected) cycles linking each target sense back to the source sense in the underlying undirected noisy graph. We did not implement the DFS in the undirected setting because it is equivalent to the Undirected Cycles; neither did we implement the undirected versions of Random Walks,

| Algorithm | PERFORMANCE | | | |
|---|---|---|---|---|
| | P | R | F1 | A |
| Undirected Cycles $1/e^l$ | 76.67 | 67.16 | 66.73 | 71.60 |
| Undirected Cycles $1/l$ | 76.56 | 67.06 | 66.63 | 71.50 |
| Undirected Cycles 1 | 76.12 | 66.67 | 66.25 | 71.08 |

Table 8: Disambiguation performance on the Ragazzini-Biagi dataset using an undirected model.

| Algorithm | PERFORMANCE WITH FS | | | |
|---|---|---|---|---|
| | P | R | F1 | A |
| Undirected Cycles $1/e^l$ | 77.50 | 78.10 | 77.50 | 77.80 |
| Undirected Cycles $1/l$ | 77.40 | 78.01 | 77.40 | 77.70 |
| Undirected Cycles 1 | 77.01 | 77.61 | 77.01 | 77.31 |

Table 9: Disambiguation performance on the Ragazzini-Biagi dataset using an undirected model (using the FS baseline).

Markov Chains and PPR, because they are broadly equivalent to Degree in an undirected setting (Upstill, Craswell, & Hawking, 2003). As shown in Table 8, Undirected Cycles yields a 66% F1 performance and 71% accuracy (almost regardless of the $\omega$ function). Consistently with our previous experiments, allowing the algorithm to resort to the FS Baseline as back-off strategy boosts performance up to 77-78% (with $\omega = 1/e^l$ producing the best results, see Table 9). Nonetheless, Undirected Cycles performs significantly worse than Cycles and CQC.

The reason for this behaviour lies in the strong disambiguation evidence provided by the directed flow of information. In fact, not accounting for directionality leads to a considerable *loss of information*, since we would be treating two different scenarios in the same way: one in which $s \rightarrow t$ and another one in which $s \leftrightarrows t$.

For example, in the directed setting two senses $s$ and $t$ which reciprocally link to one another ($s \leftrightarrows t$) create a cycle of length 2 ($s \rightarrow t \rightarrow s$); in an undirected setting, instead, the two edges are merged ($s - t$) and no supporting cycles of length 2 can be found. As a result we are not considering the fact that $t$ *translates back to* $s$, which is a precious piece of information! Furthermore an undirected cycle is likely to correspond to a noisy, illegal quasi-cycle (cf. Figure 3(f)), i.e., one which could contain any sequence whatsoever of plain and reversed edges. Consequently, in the undirected setting meaningful and nonsensical paths are lumped together.

## 6. Dictionary Enhancement

We now present an application of the CQC algorithm to the problem of enhancing the quality of a bilingual dictionary.

## 6.1 Ranking Translation Senses

As explained in Section 4, the application of the CQC algorithm to a sense entry determines, together with a sense choice, a ranking for the senses chosen for its translations. For instance, the most appropriate senses for the translations of *language* (cf. Section 4.1) are chosen on the basis of the following scores: 0.009 ($lingua_n^1$), 0.194 ($lingua_n^2$), 0.009 ($linguaggio_n^1$), 0.1265 ($linguaggio_n^2$), 0.0675 ($linguaggio_n^3$). The higher the score for the target translation, the higher the confidence in selecting the corresponding sense. In fact, a high score is a clear hint of a high amount of connectivity conveyed from the target translation back to the source sense. As a result, the following senses are chosen in our example: $lingua_n^2$, $linguaggio_n^2$. Our hunch is that this confidence information can prove to be useful not only in disambiguating dictionary translations, but also in identifying recurring problems dictionaries tend to suffer from.

For instance, assume an English word $w$ translates to an Italian word $w'$ but no sense entry of $w'$ in the bilingual dictionary translates back to $w$. An example where such a shortcoming could be fixed is the following: $wood_n^2 \to bosco$ but no sense of *bosco* translates back into *wood* (here $wood_n^2$ and *bosco* refer to the forest sense). However, this phenomenon does not always need to be solved. This might be the case if $w$ is a relevant (e.g., popular) translation for $w'$, but $w'$ is not a frequent term. For instance, $idioma_n^1$ ($idiom_n$ in english) translates to *language* and no sense of *language* has *idioma* as its translation. This is correct because we do not expect *language* to translate into such an uncommon word as *idioma*.

But how can we decide whether a problem of this kind needs to be fixed (like *bosco*) or not (like *idioma*)? To answer this question we will exploit the confidence scores output by the CQC algorithm. In fact, applying the CQC algorithm to the pair $wood_n^2$, $bosco_n^1$ we obtain a score of 0.2 (indicating that $bosco_n^1$ should point back to *wood*)[4], while on the pair $idioma_n^1$, $language_n^1$ we get a score of 0.07 (pointing out that $idioma_n^1$ is not at easy reach from $language_n^1$).

## 6.2 Patterns for Enhancing a Bilingual Dictionary

In this section, we propose a methodology to enhance the bilingual dictionary using the sense rankings provided by the CQC algorithm. In order to solve relevant problems raised by the Zanichelli lexicographers on the basis of their professional experience, we identified the following 6 issues, each characterized by a specific graph pattern:

- **Misalignment.** The first pattern is of the kind $s_w \to s_{w'} \not\to s_w$, where $s_w$ is a sense of $w$ in the source language, $s_{w'}$ is a sense of $w'$ in the target language, and $\to$ denotes a translation in the dictionary. For instance, $buy_n^1$ is translated as $compera_n^1$, but $compera_n^1$ is not translated as $buy_n^1$. A high-ranking sense such as $compera_n^1$ implies that this issue should be solved.

- **Partial alignment.** This pattern is of the kind $s_w \to s_{w'} \to s_{w''w}$ or $s_{w''w} \to s_{w'} \to s_w$ where $s_w$ and $s_{w''w}$ are senses in the source language, $w''w$ is a compound that ends with $w$, and $s_{w'}$ is a sense in the target language. For instance, $repellent_n^1$ is translated as $insettifugo_n^1$, which in turn translates to $insect\ repellent_n^1$.

---

4. Note that in practice values greater than 0.3 are very unlikely.

| issue | pattern | sense entry | example |
|---|---|---|---|
| misalignment | $s_w \leftrightarrow s_{w'}$ | $s_w$ <br> $s_{w'}$ | **buy** *n.* **1** *(fam.)* acquisto; compera. <br> **compera** *n.* **1** purchase; shopping. |
| partial alignment | $s_w \leftrightarrow s_{w'}$, $s_{w''w}$ | $s_w$ <br> $s_{w'}$ | **repellent** *n.* **1** sostanza repellente; insettifugo. <br> **insettifugo** *n.* **2** insect repellent. |
| missing lemma | $s_w \to s_{w'}$ | $s_w$ <br> $s_{w'}$ | **persistente** *a.* **1** persistent; persisting. <br> **persisting** *a.* *(not available in the dictionary).* |
| use of reference | $s_w \to s_{w'} \to s_{w''}$ | $s_w$ <br> $s_{w'}$ <br> $s_{w''}$ | **pass** *n.* **3** tesserino *(di autobus, ecc.)*. <br> **tesserino** *n.* **1** → tessera. <br> **tessera** *n.* **1** card; ticket; pass. |
| use of variant | $s_w \to s_{w'}, s_{w''}$ | $s_w$ <br> $s_{w'}, s_{w''}$ | **riscontro** *n.* **6** reply; acknowledgment. <br> **acknowledgement, acknowledgment** *n.* **3** conferma di ricevuta; riscontro. |
| inconsistent spelling | $s_w \to s_{w'}$, $s_{w''}$ | $s_w$ <br> $s_{w''}$ | **asciugacapelli** *n.* **1** hair-dryer. <br> **hair dryer** *n.* **1** asciugacapelli. |

Table 10: The set of graph patterns used for enhancement suggestions.

- **Missing lemma.** This pattern is of the kind $s_w \to s_{w'}$ where $s_{w'}$ does not exist in the dictionary. For example, $persistente_a^1$ is translated as $persistent_a$, however the latter lemma does not exist in the dictionary lexicon.

- **Use of reference.** This pattern is of the kind $s_w \to s_{w'} \to s_{w''} \to s_w$ where $s_{w'}$ is a reference to $s_{w''}$. For example, $pass_n^3$ is translated as $tesserino_n^1$, while the latter refers to $tessera_n^1$, which in turn is translated as $pass_n$. However, for clarity's sake, double referencing should be avoided within dictionaries.

- **Use of variant.** This pattern is of the kind $s_w \to s_{w'}$ and $s_{w''} \to s_w$, where $w''$ is a variant of $w'$. For example, $riscontro_n^6$ is translated as $acknowledgment_n^1$. However, this is a just variant of the main form $acknowledgement_n^1$. In the interests of consistency the main form should always be preferred.

- **Inconsistent spelling.** This pattern is of the kind $s_w \to s_{w'}$ and $s_{w''} \to s_w$ where $w$ and $w'$ differ only by minimal spelling conventions. For example, $asciugacapelli_n^1$ is translated as $hair\text{-}dryer_n^1$, while $hair\ dryer_n^1$ is translated as $asciugacapelli_n^1$. The inconsistency between *hair-dryer* and *hair dryer* must be solved in favour of the latter, which is a lemma defined within the dictionary.

Table 10 presents the above patterns in the form of graphs together with examples.

Next, we collected all the pattern occurrences in the Ragazzini-Biagi bilingual dictionary and ranked them by the CQC scores assigned to the corresponding translation in the source entry. An excerpt of the top- and bottom-ranking issues for the misalignment pattern is reported in Table 11.

| source sense | target sense | CQC score |
|:---:|:---:|:---:|
| $still_a^{1.A.1}$ | $immobile_a^{A.2}$ | 0.3300 |
| $burrasca_n^1$ | $storm_n^1$ | 0.3200 |
| $achievement_n^2$ | $impresa_n^2$ | 0.3100 |
| $\vdots$ | $\vdots$ | $\vdots$ |
| $phat_a^1$ | $grande_a^{A.1}$ | 0.0001 |
| $opera_n^5$ | $society_n^2$ | 0.0001 |

Table 11: Top- and bottom-ranking dictionary issues identified using the misalignment pattern.

| issue | % accepted | no. absolute |
|:---|:---:|---:|
| misalignment | 80.0 | 118904 |
| partial alignment | 40.0 | 8433 |
| missing lemma | 21.0 | 15955 |
| use of reference | 84.5 | 167 |
| use of variant | 83.5 | 1123 |
| inconsistent spelling | 98.0 | 12 |

Table 12: Enhancement suggestions accepted.

## 7. Evaluation: Dictionary Enhancement

In the following two subsections we describe the experimental setup and give the results of our dictionary enhancement experiment.

### 7.1 Experimental Setup

The aim of our evaluation is to show that the higher the confidence score the higher the importance of the issue for an expert lexicographer. Given such an issue (e.g., misalignment), we foresee two possible actions to be taken by a lexicographer: "apply some change to the dictionary entry" or "ignore the issue". In order to assess the quality of the issues, we prepared a dataset of 200 randomly-sampled instances for each kind of dictionary issue (i.e., 200 misalignments, 200 uses of variants, etc.). Overall the dataset included 1,200 issue instances (i.e., $200 \cdot 6$ issue types). The dataset was manually annotated by expert lexicographers, who decided for each issue whether a change in the dictionary was needed (positive response) or not (negative response). Random sampling guarantees that the dataset has a distribution comparable to that of the entire set of instances for an issue of interest.

suggestions accepted / suggestions presented

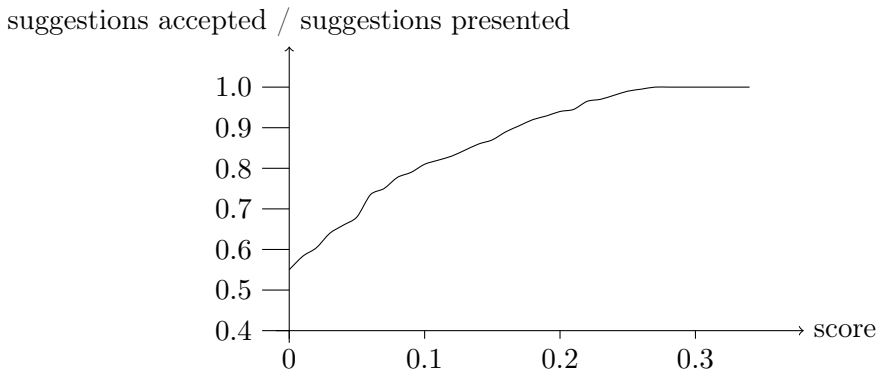

Figure 6: Performance trend of enhancement suggestions accepted by score for the misalignment issue.

## 7.2 Results

We report the results for each issue type in Table 12. We observed an acceptance percentage ranging between 80.0 and 84.5% for three of the issues, namely: misalignment, use of reference and use of variant, thus indicating a high level of reliability for the degree of importance calculated for these issues. We note however that semantics cannot be of much help in the case of missing lemmas, partial alignment and inconsistent spelling. In fact these issues inevitably cause the graphs to be disconnected and thus the disambiguation scores equal 0.

To determine whether our score-based ranking impacts the degree of reliability of our enhancement suggestions we graphed the percentage of accepted suggestions by score for the misalignment issue (Figure 6). As expected, the higher the disambiguation score, the higher the percentage of suggestions accepted by lexicographers, up to 99% when the score > 0.27. We observed similar trends for the other issues.

## 8. Synonym Extraction

In the previous sections, we have shown how to use cycles and quasi-cycles to extend bilingual dictionary entries with sense information and tackle important dictionary issues. We now propose a third application of the CQC algorithm to enrich the bilingual dictionary with synonyms, a task referred to as synonym extraction. The task consists of automatically identifying appropriate synonyms for a given lemma. Many efforts have been made to develop automated methods that collect synonyms. Current approaches typically rely either on statistical methods based on large corpora or on fully-fledged semantic networks such as WordNet (a survey of the literature in the field is given in Section 10). Our approach is closer to the latter direction, but relies on a bilingual machine readable dictionary (i.e., a resource with no explicit semantic relations), rather than a full computational lexicon. We exploit cross-language connections to identify the most appropriate synonyms for a given word using cycles and quasi-cycles.

The idea behind our synonym extraction approach is as follows: starting from some node(s) in the graph associated with a given word, we perform a cycle and quasi-cycle search (cf. Section 4). The words encountered in the cycles or quasi-cycles are likely to be closely related to the word sense we started from and they tend to represent good synonym candidates in the two languages. We adopted two synonym extraction strategies:

- *sense-level* synonym extraction: the aim of this task is to find synonyms of a given *sense s* of a word $w$.

- *word-level* synonym extraction: given a word $w$, we collect the union of the synonyms for all senses of $w$.

In both cases we apply CQC to obtain a set of paths $P$ (respectively starting from a given sense $s$ of $w$ or from any sense of $w$). Next, we rank each candidate synonym according to the following formula:

$$score(w') = \sum_{p \in P(w')} \frac{1}{e^{length(p)}} \tag{5}$$

which provides a score for a synonym candidate $w'$, where $P(w')$ is the set of (quasi-)cycles passing through a sense of $w'$. In the sense-level strategy $P(w')$ contains all the paths starting from sense $s$ of our source word $w$, whereas in the word-level strategy $P(w')$ contains paths starting from any sense of $w$. In contrast to other approaches in the literature, our synonym extraction approach actually produces not only synonyms, but also their senses according to the dictionary sense inventory. Further, thanks to the above formula, we are able to rank synonym senses from most to less likely. For example, given the English sense $capable_a^1$, the system outputs the following ordered list of senses:

| Lang. | Word sense | Score |
|-------|------------|-------|
| it | $abile_a^1$ | 13.34 |
| it | $capace_a^2$ | 8.50 |
| en | $able_a^1$ | 4.42 |
| en | $skilful_a^1$ | 3.21 |
| it | $esperto_a^{A.1}$ | 3.03 |
| en | $clever_a^1$ | 2.61 |
| en | $deft_a^1$ | 1.00 |
| $\vdots$ | $\vdots$ | $\vdots$ |
| en | $crafty_a^1$ | 0.18 |
| it | $destro_a^{A.1}$ | 0.17 |

In the word-level strategy, instead, synonyms are found by performing a CQC search *starting from each sense* of the word $w$, and thus collecting the union of all the paths obtained in each individual visit. As a result we can output the list of all the *words* which are likely to be synonym candidates. For example, given the English word *capable*, the system outputs the following ordered list of words:

| Lang. | Word | Score |
|:---:|:---:|:---:|
| it | *abile* | 12.00 |
| it | *capace* | 10.65 |
| en | *clever* | 3.45 |
| en | *able* | 2.90 |
| it | *esperto* | 2.41 |
| en | *skilful* | 2.30 |
| en | *deft* | 0.54 |
| ⋮ | ⋮ | ⋮ |
| it | *sapiente* | 0.18 |
| en | *expert* | 0.16 |

## 9. Evaluation: Synonym Extraction

We now describe the experimental setup and discuss the results of our synonym extraction experiment.

### 9.1 Experimental Setup

#### 9.1.1 Dataset

To compare the performance of CQC on synonym extraction with existing approaches, we used the Test of English as a Foreign Language (TOEFL) dataset provided by ETS via Thomas Landauer and coming originally from the Educational Testing Service (Landauer & Dumais, 1997). This dataset is part of the well-known TOEFL test used to evaluate the ability of an individual to use and understand English. The dataset includes 80 question items, each presenting:

1. a sentence where a target word $w$ is emphasized;

2. 4 words listed as possible synonyms for $w$.

The examinee is asked to indicate which one, among the four presented choices, is more likely to be the right synonym for the given word $w$. The examinee's language ability is then estimated to be the fraction of correct answers. The performance of automated systems can be assessed in the very same way.

#### 9.1.2 Algorithms

We performed our experiments with the same algorithms used in Section 5.1.4 and compared their results against the best ones known in the literature. All of our methods are based on some sort of graph path or cycle collection. In order to select the best synonym for a target word, we used the approach described in Section 8 for all methods but Markov chains and PPR. For the latter we replaced equation 5 with the corresponding scoring function of the method (cf. Section 5.1.4). We also compared with the best approaches for synonym extraction in the literature, including:

- **Product Rule (PR)** (Turney, Littman, Bigham, & Shnayder, 2003): this method – which achieves the highest performance – combines various different modules. Each module produces a probability distribution based on a word closeness coefficient calculated on the possible answers the system can output and a merge rule is then applied to integrate all four distributions into a single one.

- **Singular Value Decomposition (LSA)** (Rapp, 2003), an automatic Word Sense Induction method which aims at finding sense descriptors for the different senses of ambiguous words. Given a word, the twenty most similar words are considered good candidate descriptors. Then pairs are formed and classified according to two criteria: i) two words in a couple should be as dissimilar as possible; ii) their cooccurrence vectors should sum to the ambiguous word cooccurrence vector (scaled by 2). Finally, words with the highest score are selected.

- **Generalized Latent Semantic Analysis (GLSA)** (Matveeva, Levow, Farahat, & Royer, 2005), a corpus-based method which builds term-vectors and represents the document space in terms of vectors. By means of Singular Value Decomposition and Latent Semantic Analysis they obtain the similarity matrix between the words of a prefixed vocabulary and extract the related document matrix. Next, synonyms of a word are selected on the basis of the highest cosine-similarity between the candidate synonym and the fixed word.

- **Positive PMI Cosine (PPMIC)** (Bullinaria & Levy, 2007) systematically explores several possibilities of representation for the word meanings in the space of cooccurrence vectors, studying and comparing different information metrics and implementation details (such as the cooccurrence window or the corpus size).

- **Context-window overlapping (CWO)** (Ruiz-Casado, M., E., & Castells, 2005) is an approach based on the key idea that synonymous words can be replaced in most contexts. Given two words, their similarity is measured as the number of contexts that can be found by replacing each word with the other, where the context is restricted to an $L$-window of open-class words in a Google snippet.

- **Document Retrieval PMI (PMI-IR)** (Terra & Clarke, 2003) integrates many different word similarity measures and cooccurrence estimates. Using a large corpus of Web data they analyze how the corpus size influences the measure performance and compare a window- with a document-oriented approach.

- **Roget's Thesaurus system (JS)** (Jarmasz & Szpakowicz, 2003), exploits Roget's thesaurus taxonomy and WordNet to measure semantic similarity. Given two words their closeness is defined as the minimum distance between the nodes associated with the words. This work is closest to our own in that the structure of knowledge resources is exploited to extract synonyms.

## 9.2 Results

In Table 13 and 14 we report the performance (precision and recall, respectively) of our algorithms on the TOEFL with maximum path length $\delta$ varying from 2 to 6. The best results are obtained for all algorithms (except for Markov chains) when $\delta = 6$, as this value makes it easier to find near synonyms that cannot be immediately obtained as translations of the target word in the dictionary. We attribute the higher recall (but lower precision) of Markov chains to the amount of noise accumulated after only few steps. Interestingly, PPR (which is independent of parameter $\delta$, and therefore is not shown in Tables 13 and 14) obtained comparable performance, i.e., 94.55% precision and 65% recall. Thus, CQC is the best graph-based approach achieving 93% precision and 85% recall. This result corroborates our previous findings (cf. Section 5).

| Algorithm | Maximum path length | | | | |
|---|---|---|---|---|---|
| | 2 | 3 | 4 | 5 | 6 |
| CQC | 100.00 | 97.56 | 98.15 | 98.36 | 93.15 |
| Cycles | 100.00 | 97.50 | 97.83 | 98.00 | 96.43 |
| DFS | 97.67 | 97.78 | 97.82 | 98.04 | 96.36 |
| Random walks | 97.44 | 95.12 | 97.56 | 97.62 | 97.83 |
| Markov chains | 94.91 | 86.76 | 85.29 | 85.29 | 85.29 |

Table 13: Precision of our graph-based algorithms on the TOEFL dataset.

| Algorithm | Maximum path length | | | | |
|---|---|---|---|---|---|
| | 2 | 3 | 4 | 5 | 6 |
| CQC | 47.50 | 50.00 | 66.25 | 75.00 | 85.00 |
| Cycles | 47.50 | 48.75 | 56.25 | 61.25 | 67.50 |
| DFS | 52.50 | 55.00 | 56.25 | 62.50 | 66.25 |
| Random walks | 47.50 | 48.75 | 50.00 | 51.25 | 56.25 |
| Markov chains | 70.00 | 73.75 | 72.50 | 72.50 | 72.50 |

Table 14: Recall of our graph-based algorithms on the TOEFL dataset.

Table 15 shows the results of the best systems in the literature and compares them with CQC.[5] We note that the systems performing better than CQC exploit a large amount of information: for example Rapp (2003) uses a corpus of more than 100 million words of everyday written and spoken language, while Matveeva et al. (2005) draw on more than 1 million New York Times articles with a 'history' label. Even if they do not rely on a manually-created lexicon, they do have to cope with the extremely high term-space dimension and need to adopt some method to reduce dimensionality (i.e., either using Latent Semantic Indexing on the term space or reducing the vocabulary size according to some general strategy such as selecting the top frequent words).

It is easy to see how our work stands above all lexicon-based ones, raising performance from 78.75% up to 85% recall. In Table 15 we also report the performance of other lexicon-based approaches in the literature (Hirst & St-Onge, 1998; Leacock & Chodorow, 1998; Jarmasz & Szpakowicz, 2003). We note that our system exploits the concise edition of the Ragazzini bilingual dictionary which omits lots of translations (i.e., edges) and senses which are to be found in the complete edition of the dictionary. Our graph algorithm could readily take advantage of the richer structure of the complete edition to achieve even better performance.

---

5. Further information about the state of the art for the TOEFL test can be found at the following web site: `http://aclweb.org/aclwiki/index.php?title=TOEFL_Synonym_Questions_(State_of_the_art)`

| Algorithm | Author(s) / Method | Resource type | Recall (%) |
|---|---|:---:|:---:|
| PR | Turney et al. (2003) | Hybrid | 97.50 |
| LSA | Rapp (2003) | Corpus-based | 92.50 |
| GLSA | Matveeva et al. (2005) | Corpus-based | 86.25 |
| **CQC** | **Flati and Navigli (2012)** | **Lexicon-based** | **85.00** |
| PPMIC | Bullinaria and Levy (2007) | Corpus-based | 85.00 |
| CWO | Ruiz-Casado et al. (2005) | Web-based | 82.55 |
| PMI-IR | Terra and Clarke (2003) | Corpus-based | 81.25 |
| JS | Jarmasz and Szpakowicz (2003) | Lexicon-based | 78.75 |
| HSO | Hirst and St-Onge (1998) | Lexicon-based | 77.91 |
| PairClass | Turney (2008) | Corpus-based | 76.25 |
| DS | Pado and Lapata (2007) | Corpus-based | 73.00 |
| Human | Average non-English US college applicant | Human | 64.50 |
| Random | Random guessing | Random | 25.00 |
| LC | Leacock and Chodorow (1998) | Lexicon-based | 21.88 |

Table 15: Recall of synonym extraction systems on the TOEFL dataset.

Another interesting aspect is the ability of CQC to rank synonym candidates. To better understand this phenomenon, we performed a second experiment. We selected 100 senses (50 for each language). We applied the CQC algorithm to each of them and also to their lemmas. In the former case a sense-tagged list was returned; in the latter the list contained just words. Then we determined the precision of CQC in retrieving the top ranking K synonyms (precision@K) according to the algorithm's score. We performed our evaluation at both the sense- and the word-level, as explained in Section 8. In Table 16 we report the precision@K calculated at both levels when $K = 1, \ldots, 10$. Note that, when K is sufficiently small ($K \leq 4$), the sense-level extraction achieves performance similar to the word-level one, while being semantically precise. However, we observe that with larger values of K the performance difference increases considerably.

## 10. Related Work

We now review the literature in the three main fields we have dealt with in this paper, namely: gloss disambiguation (Section 10.1), dictionary enhancement (Section 10.2) and synonym extraction (Section 10.3).

### 10.1 Gloss Disambiguation

Since the late 1970s much work on the analysis and disambiguation of dictionary glosses has been done. This includes methods for the automatic extraction of taxonomies from lexical resources (Litkowski, 1978; Amsler, 1980), the identification of genus terms (Chodorow, Byrd, & Heidorn, 1985) and, more in general, the extraction of explicit information from machine-readable dictionaries (see, e.g., Nakamura & Nagao, 1988; Ide & Véronis, 1993), as well as the construction of ambiguous semantic networks from glosses (Kozima & Furugori, 1993). A relevant project in this direction is MindNet (Vanderwende, 1996; Richardson, Dolan, & Vanderwende, 1998), a lexical knowledge base obtained from the automated extraction of lexico-semantic information from two machine-readable dictionaries.

| Level | K | Precision | Correct/Given |
|-------|---|-----------|---------------|
| Sense | 1 | 79.00 | 79 / 100 |
|       | 2 | 75.50 | 151 / 200 |
|       | 3 | 70.33 | 211 / 300 |
|       | 4 | 67.01 | 266 / 397 |
|       | 5 | 63.41 | 312 / 492 |
|       | 6 | 60.31 | 354 / 587 |
|       | 7 | 59.24 | 404 / 682 |
|       | 8 | 57.01 | 443 / 777 |
|       | 9 | 55.28 | 482 / 872 |
|       | 10 | 53.06 | 512 / 965 |
| Word  | 1 | 80.00 | 80 / 100 |
|       | 2 | 74.50 | 149 / 200 |
|       | 3 | 71.67 | 215 / 300 |
|       | 4 | 72.50 | 290 / 400 |
|       | 5 | 71.20 | 356 / 500 |
|       | 6 | 68.83 | 413 / 600 |
|       | 7 | 67.29 | 471 / 700 |
|       | 8 | 65.88 | 527 / 800 |
|       | 9 | 64.22 | 578 / 900 |
|       | 10 | 62.90 | 629 /1000 |

Table 16: Precision@K of the CQC algorithm in the sense and word synonym extraction task.

More recently, a set of heuristics has been proposed to semantically annotate WordNet glosses, leading to the release of the eXtended WordNet (Harabagiu et al., 1999; Moldovan & Novischi, 2004). Among the heuristics, the cross reference heuristic is the closest technique to our notion of (quasi-)cyclic patterns. Given a pair of words $w$ and $w'$, this heuristic is based on the occurrence of $w$ in the gloss of a sense $s'$ of $w'$ and, vice versa, of $w'$ in the gloss of a sense $s$ of $w$. In other words, a cycle $s \rightarrow s' \rightarrow s$ of length 2 is sought. Recently, a similar consideration has been put forward proposing that probabilistic translation circuits can be used as evidence to automatically acquire a multilingual dictionary (Mausam et al., 2009).

Based on the eXtended WordNet, a gloss disambiguation task was organized at Senseval-3 (Litkowski, 2004). Most notably, the best performing systems, namely the TALP system (Castillo et al., 2004), and SSI (Navigli & Velardi, 2005), are knowledge-based and rely on rich knowledge resources: respectively, the Multilingual Central Repository (Atserias, Vil-larejo, Rigau, Agirre, Carroll, Magnini, & Vossen, 2004), and a proprietary lexical knowledge base (cf. Navigli & Lapata, 2010).

However, the literature in the field of gloss disambiguation is focused only on monolingual dictionaries, such as WordNet and LDOCE. To our knowledge, CQC is the first algorithm aimed at disambiguating the entries of a bilingual dictionary: our key idea is to harvest (quasi-)cyclic paths from the dictionary – viewed as a noisy graph – and use them to associate meanings with the target translations. Moreover, in contrast to many disambiguation methods in the literature (Navigli, 2009b), our approach works on bilingual machine-readable dictionaries and does not exploit lexical and semantic relations, such as those available in computational lexicons like WordNet.

## 10.2 Dictionary Enhancement

The issue of improving the quality of machine-readable dictionaries with computational methods has been poorly investigated so far. Ide and Véronis (1993, 1994), among others, have been working on the identification of relevant issues when transforming a machine-readable dictionary into a computational lexicon. These include overgenerality (e.g., a newspaper defined as an artifact, rather than a publication), inconsistent definitions (e.g., two concepts defined in terms of each other), meta-information labels and sense divisions (e.g., fine-grained vs. coarse-grained distinctions). Only little work has been done on the automatic improvement of monolingual dictionaries (Navigli, 2008), as well as bilingual resources, for which a gloss rewriting algorithm has been proposed (Bond, Nichols, & Breen, 2007). However, to our knowledge, the structure of bilingual dictionaries has never previously been exploited for the purpose of suggesting dictionary enhancements. Moreover, we rank our suggestions on the basis of semantic-driven confidence scores, thus submitting to the lexicographer more pressing issues first.

## 10.3 Synonym Extraction

Another task aimed at improving machine-readable dictionaries is that of synonym extraction. Many efforts have been made to automatically collect a set of synonyms for a word of interest. We introduced various methods aimed at this task in Section 8. Here we distinguish in greater detail between corpus-based (i.e., statistical) and lexicon-based (or knowledge-based) approaches.

Corpus-based approaches typically harvest statistical information about word occurrences from large corpora, inferring probabilistic clauses such as "word $w$ is likely to appear (i.e., cooccur) together with word $y$ with probability $p$". Thus, word similarity is approximated with word distance functions. One common goal is to build a cooccurrence matrix; this can be done directly via corpus analysis or indirectly by obtaining its vector space representation.

The most widespread statistical method (Turney et al., 2003; Bullinaria & Levy, 2007; Ruiz-Casado et al., 2005; Terra & Clarke, 2003) is to estimate the word distance by counting the number of times that two words appear together in a corpus within a fixed $k$-sized window, followed by a convenient normalization. This approach suffers from the well-known data sparseness problem; furthermore it introduces the additional window-size parameter $k$ whose value has to be tuned.

A similar statistical approach consists of building a vocabulary of terms $V$ from a corpus $C$ and then representing a document by means of the elements of $V$ contained therein. In this framework a document is represented as a vector, a corpus as a term-document matrix $L$ as well as a document-term matrix $L'$. The matrix product $LL'$ represents the cooccurrence matrix which gives a measure of word closeness.

For computational reasons, however, it is often desirable to shrink the vocabulary size. Classical algebraic methods, such as Singular Value Decomposition (SVD), can be applied to synonym extraction (Rapp, 2003; Matveeva et al., 2005), because they are able to produce a smaller vocabulary $V'$ representing the concept space. These methods do not take into account the relative word position, but only cooccurrences within the same document, so less information is usually considered. On the other hand, by virtue of SVD, a more significant concept space is built and documents can be more suitably represented.

Lexicon-based approaches (Jarmasz & Szpakowicz, 2003; Blondel & Senellart, 2002) are an alternative to purely statistical ones. Graph models are employed in which words are represented by nodes and relations between words by edges between nodes. In this setting, no corpus is required. Instead two words are deemed to be synonyms if the linking path, if any, satisfies some structural criterion, based on length, structure or connectivity degree. Our application of CQC to the synonym extraction problem follows this direction. However, in contrast to existing work in the literature, we do not exploit any lexical or semantic relation between concepts, such as those in WordNet, nor any lexical pattern as done by Wang and Hirst (2012). Further, we view synonym extraction as a dictionary enrichment task that we can perform at a bilingual level.

## 11. Conclusions

In this paper we presented a novel algorithm, called Cycles and Quasi-Cycles (CQC), for the disambiguation of bilingual machine-readable dictionaries. The algorithm is based on the identification of (quasi-)cycles in the noisy dictionary graph, i.e., circular edge sequences (possibly with some consecutive edges reversed) relating a source word sense to a target one.

The contribution of the paper is threefold:

1. We show that our notion of (quasi-)cyclic patterns enables state-of-the-art performance to be attained in the disambiguation of dictionary entries, surpassing all other disambiguation approaches (including the popular PPR), as well as a competitive baseline such as the first sense heuristic. Crucially, the introduction of reversed edges allows us to find more semantic connections, thus substantially increasing recall while keeping precision very high.

2. We explore the novel task of dictionary enhancement by introducing graph patterns for a variety of dictionary issues, which we tackle effectively by means of the CQC algorithm. We use CQC to rank the issues based on the disambiguation score and present enhancement suggestions automatically. Our experiments show that the higher the score the more relevant the suggestion. As a result, important idiosyncrasies such as missing or redundant translations can be submitted to expert lexicographers, who can review them in order to improve the bilingual dictionary.

3. We successfully apply CQC to the task of synonym extraction. While data-intensive approaches achieve better performance, CQC obtains the best result among lexicon-based systems. As an interesting side effect, our algorithm produces sense-tagged synonyms for the two languages of interest, whereas state-of-the-art approaches all focus on a single language and do not produce sense annotations for synonyms.

The strength of our approach lies in its weakly supervised nature: the CQC algorithm relies exclusively on the structure of the input bilingual dictionary. Unlike other research directions, no further resource (such as labeled corpora or knowledge bases) is required.

The paths output by our algorithm for the dataset presented in Section 5.1 are available from `http://lcl.uniroma1.it/cqc`. We are scheduling the release of a software package which allows for the application of the CQC algorithm to any resource for which a standard interface can be implemented.

As regards future work, we foresee several developments of the CQC algorithm and its applications: starting from the work of Budanitsky and Hirst (2006), we plan to experiment with cycles and quasi-cycles when used as a semantic similarity measure, and compare them with the most successful existing approaches. Moreover, although in this paper we focused on the disambiguation of dictionary glosses, exactly the same approach can be applied to the disambiguation of collocations using any dictionary of choice (along the lines of Navigli, 2005), thus providing a way of further enriching lexical knowledge resources with external knowledge.

## Acknowledgments

The authors gratefully acknowledge the support of the ERC Starting Grant MultiJEDI No. 259234. The authors wish to thank Jim McManus, Simon Bartels and the three anonymous reviewers for their useful comments on the paper, and Zanichelli for making the Ragazzini-Biagi dictionary available for research purposes.